\documentclass{article} 
\usepackage{iclr2026_conference,times}

\usepackage{amsmath,amsfonts,bm}

\def\eqref#1{equation~\ref{#1}}

\def\1{\bm{1}}

\DeclareMathAlphabet{\mathsfit}{\encodingdefault}{\sfdefault}{m}{sl}
\SetMathAlphabet{\mathsfit}{bold}{\encodingdefault}{\sfdefault}{bx}{n}

\usepackage{hyperref}
\usepackage{url}
\usepackage{graphicx}
\usepackage{wrapfig}
\usepackage{booktabs}
\usepackage{multirow}

\title{PE-EK-PINN: Physics Embedding with Evolving Kernel for Scalable Physics-Informed Neural Networks}

\author{{Huiwen Zhang, Feng Ye, and Chu Ma } \\
Department of Electrical \& Computer Engineering\\
Madison, WI, 53706, USA \\
\texttt{\{hzhang2279, feng.ye, chu.ma\}@wisc.edu} 
}

\iclrfinalcopy 
\begin{document}

\maketitle

\begin{abstract}
Physics-Informed Neural Networks (PINNs) embed governing equations into deep learning, but enforce them only through loss residuals, leaving highly oscillatory wave behavior to be discovered by optimization. As a result, methods that achieve relative $L_2$ errors below $10^{-3}$ on standard manufactured Helmholtz benchmarks can fail on practical radiation problems involving singular excitations, absorbing boundaries, and wave fields spanning tens of wavelengths. Architectural physics embedding addresses this limitation by factorizing the field into analytically derived oscillatory kernels and learnable envelopes. However, the kernel dictionary must be manually constructed and scales with the number of elementary units, growing exponentially with the depth of hierarchically structured systems such as antenna arrays and metasurfaces. We propose PE-EK-PINN (Physics Embedded with Evolving Kernels), which treats physics kernels as reusable learned representations rather than fixed analytical inputs. A converged subsystem field is frozen and promoted to an evolved kernel, whose transformed copies are reused to represent higher-level configurations without deriving new governing equations. The resulting hierarchy makes the peak number of active kernels independent of system size and reduces cumulative training cost from $\mathcal{O}(N)$ to $\mathcal{O}(\log N)$. Experiments on dipole arrays, composite line-source geometries, and cross arrays demonstrate the dramatic training cost reduction, while achieving a reduced or comparable relative $L_2$ error. One notable example is PE-EK-PINN solves a $256$-dipole array more than 30 times faster than direct PE-PINN.
\end{abstract}

\section{Introduction}

Physics-Informed Neural Networks (PINNs) \citep{raissi2019pinn} solve partial differential equations (PDEs) by embedding governing equations, boundary conditions, and initial conditions directly into the training objective, providing a mesh-free framework for scientific computing \citep{karniadakis2021piml,cuomo2022survey}. While broadly applicable, PINNs rely on optimization alone to discover the solution. As a result, they inherit the well-known spectral bias of neural networks \citep{rahaman2019spectral,xu2020frequency}: low-frequency components are learned significantly faster than high-frequency ones, making oscillatory wave fields particularly challenging to represent accurately.
This limitation is often obscured by standard benchmarks. On the manufactured Helmholtz problems commonly used in the PINN literature, recent methods achieve relative $L_2$ errors below $10^{-3}$. Yet on a physically realistic dipole radiating at $2.4$~GHz, the same methods can fail dramatically. For example, CoPINN \citep{duan2025copinn} degrades from $5.0\times10^{-3}$ on the manufactured benchmark to $9.95\times10^{-1}$ on the dipole problem (Table~\ref{tab:helmholtz_comparison}), indicating almost no correspondence with the true field. The gap is not incidental. Manufactured solutions are smooth, exactly separable, and span only a few wavelengths, whereas practical radiation fields are generally non-separable and involve singular sources, absorbing boundaries, and domains extending over tens of wavelengths.

PE-PINN \citep{zhang2026physicsinformedneuralnetworksarchitectural} addresses this challenge by embedding wave physics directly into the architecture rather than relying solely on the training loss. It factorizes the field into analytically derived oscillatory kernels and learnable envelopes, allowing the network to model only the remaining smooth variation. This strategy reduces the impact of spectral bias and achieves a relative $L_2$ error of $2.14\times10^{-2}$ on the same dipole problem. However, PE-PINN introduces a new scalability bottleneck, because its kernel dictionary must be designed manually, and the number of required kernels grows with the size of the source configuration. If each elementary unit requires $q$ kernels, then a system of $N$ sources requires $qN$ kernels. In structured systems such as dipole arrays, phased arrays, and metasurfaces, $N$ grows geometrically with hierarchy depth, making kernel specification increasingly expensive. We refer to this limitation as the \emph{kernel scalability problem}.

This paper addresses that problem through a new hierarchical framework PE-EK-PINN, i.e., Physics Embedded with Evolving Kernels. In this new framework, a PE-PINN trained on a subsystem produces a converged field representation, which is then frozen and promoted to an \emph{evolved kernel}. The evolved kernel serves the same architectural role as an analytic primitive while encoding the collective wave behavior of an entire subsystem. Transformed copies of the evolved kernel are reused to construct the next-level configuration, requiring only new envelopes to be trained. Because the Helmholtz operator is invariant under rigid motions in homogeneous media, this reuse preserves physical consistency and can be applied recursively, enabling increasingly complex structures to be built from previously learned components rather than expanded into primitive kernels.

The main contributions of this work are:
\begin{itemize} 
\item We demonstrate that strong performance on manufactured Helmholtz benchmarks does not necessarily transfer to practical radiation problems, and identify manual kernel construction as the primary scalability bottleneck once architectural physics embedding is introduced.

\item We propose PE-EK-PINN, a hierarchical framework that derives reusable higher-level physics kernels from previously learned solutions, eliminating the need to formulate new analytical kernels for larger configurations.

\item We derive and experimentally validate the resulting complexity reduction. Specifically, the peak number of active kernels becomes independent of system size, while cumulative training cost scales as $\mathcal{O}(\log N)$ rather than $\mathcal{O}(N)$.

\item We evaluate the method on dipole arrays, composite line-source geometries, and cross arrays. Notably, we manage to train a 256-dipole array in 2~h~11~m compared with an extrapolated 71~h for direct PE-PINN with a fourfold reduction in relative $L_2$ error.
\end{itemize}

Throughout the paper, PE-PINN serves as the primary baseline because it already outperforms conventional PINN variants, including SPINN and CoPINN, on representative wave benchmarks (Table~\ref{tab:helmholtz_comparison}). The large-scale configurations considered here also exceed the practical capacity of those earlier methods on our single-GPU platform. Neural operators provide an alternative paradigm for PDE solving by learning mappings from problem parameters to solution fields~\cite{articledeeponet,li2021fourierneuraloperatorparametric}. While generally more scalable than PINNs for repeated evaluations, they require large amounts of high-fidelity training data, which can be prohibitively expensive to generate for the large-scale wave simulations considered in this work. Moreover, neural operators face challenges similar to those of PINNs when modeling highly oscillatory wavefields and singularities. Future work will explore using solutions generated by the proposed framework as training data for neural operators and extending the proposed kernel factorization strategy to neural operator architectures to improve convergence and the representation of wave phenomena.

\begin{table}[t]

    \centering

    \caption{
    Relative $L_2$ errors on two Helmholtz settings. The manufactured-solution block follows the protocol of prior work: $\Omega=[-1,1]^3$, $u=\sin(4\pi x_1)\sin(4\pi x_2)\sin(3\pi x_3)$, with the corresponding source term obtained by substituting $u$ into the Helmholtz equation, and the number of collocation points $N_c=32^3$. The dipole block is a $2.4$~GHz radiation problem with a singular excitation and absorbing truncation, with both methods run under identical settings. First-block baselines are as reported in~\cite{duan2025copinn}; all remaining values are from our own runs using the authors' recommended settings. 
    }
    \label{tab:helmholtz_comparison}
    \renewcommand{\arraystretch}{1.15}
    \setlength{\tabcolsep}{7pt}

    \begin{tabular}{c|l|c|c}
        \hline
        \textbf{Scenario}
        & \textbf{Method}
        & \textbf{Ref.}
        & \textbf{Rel. $L_2$} \\
        \hline

        \multirow{9}{*}{\shortstack{Manufactured Solution\\(separable, $\sim$4 wavelengths)}}
        & PINN      & \cite{raissi2019pinn}      & 0.97570 \\
        & gPINN     & \cite{yu2022gradient}      & 0.32550 \\
        & AHD-PINN  & \cite{dashtbayaz2024physics} & 0.19030 \\
        & SPINN     & \cite{cho2023spinn}        & 0.08090 \\
        & SPINN (m) & \cite{cho2023spinn}        & 0.05950 \\
        & RoPINN    & \cite{wu2024ropinn}        & 0.33380 \\
        & FPINN     & \cite{wu2025deep}          & 0.35020 \\
        & \textit{CoPINN}    & \cite{duan2025copinn}      & \textit{0.00500} \\
        & \textbf{PE-PINN} & \cite{zhang2026physicsinformedneuralnetworksarchitectural} & \textbf{0.00006} \\
        \hline

        \multirow{2}{*}{\shortstack{2.4 GHz Dipole Radiation\\(non-separable, $\sim$40 wavelengths)}}
        & \textit{CoPINN}    & \cite{duan2025copinn}      & \textit{0.99458} \\
        & \textbf{PE-PINN} & \cite{zhang2026physicsinformedneuralnetworksarchitectural} & \textbf{0.02136} \\
        \hline

    \end{tabular}
    
\vspace{-4mm}

\end{table}

\section{Related Work} \label{sec:related}

PINNs \citep{raissi2019pinn} embed governing PDEs and physical constraints into the training objective, providing a mesh-free framework for scientific machine learning \citep{karniadakis2021piml,cuomo2022survey}. Their performance is limited by the cost of dense collocation sampling, spectral bias against highly oscillatory solutions \citep{rahaman2019spectral,xu2020frequency}, and the difficult optimization of coupled PDE, boundary, and initial-condition losses \citep{wang2021gradientflow,krishnapriyan2021failure}. Existing remedies target these challenges from different directions: separable representations reduce computational cost through coordinate factorization \citep{cho2023spinn}; domain decomposition localizes learning over large domains \citep{jagtap2020xpinn,moseley2023fbpinn}; adaptive weighting and sampling improve optimization conditioning \citep{mcclenny2023sapinn,duan2025copinn}; and Fourier features and periodic activations enrich the representational basis \citep{tancik2020fourier,sitzmann2020siren}. While effective, these methods primarily improve optimization or generic function approximation. None embeds knowledge of the wave physics itself, leaving oscillatory field structure to be learned from coordinates alone.


A related issue is that much of the existing Helmholtz literature is evaluated on manufactured solutions. Common benchmarks prescribe a smooth separable field on a simple domain and derive the forcing term by substituting $u$ into the Helmholtz equation, as in SPINN \citep{cho2023spinn}, CoPINN \citep{duan2025copinn}, and related work \citep{yu2022gradient,dashtbayaz2024physics,wu2024ropinn,wu2025deep}. Such settings are useful for controlled comparison, but differ substantially from radiation problems involving singular sources, absorbing boundaries, and propagation over many wavelengths. As illustrated in Table~\ref{tab:helmholtz_comparison}, performance on the manufactured benchmark therefore does not necessarily transfer to the practical radiation setting considered in this work.

PE-PINN \citep{zhang2026physicsinformedneuralnetworksarchitectural} addresses this gap by embedding wave physics directly into the network architecture. It factorizes the field into analytically derived oscillatory kernels $\Psi_m(\mathbf{x})$ (plane-wave and spherical-wave modes) determined by the governing equations, source lo  cations, and Snell's law, modulated by learnable envelopes $A_m(\mathbf{x})$. Because the kernels carry the rapid phase variation, the network learns only a smooth residual field, effectively bypassing rather than merely mitigating spectral bias. Unlike basis-enrichment approaches, the kernels are derived from the specific physics of the problem rather than from a generic functional dictionary. Combined with incident/scattered-field separation and material-aware domain decomposition, PE-PINN handles singular sources, absorbing boundaries, and heterogeneous media directly, enabling convergence on room-scale electromagnetic problems where conventional PINNs fail within practical training budgets. However, PE-PINN still relies on manually constructed kernels whose size grows with the source configuration. This kernel scalability problem is the focus of the present work.

\section{Preliminaries and Scalability Challenge}\label{sec:preliminaries}

\subsection{Preliminaries }

For clarity, we consider a two-dimensional complex electric field governed by the homogeneous Helmholtz equation 
\begin{equation} 
E_z(\mathbf{x}) = E_z^{\mathrm{re}}(\mathbf{x}) + jE_z^{\mathrm{im}}(\mathbf{x});  \qquad \nabla^2 E_z(\mathbf{x}) + k^2 E_z(\mathbf{x}) = 0; \qquad \mathbf{x}=(x,y)\in\Omega , 
\label{eq:helmholtz} 
\end{equation} 
where $k=2\pi/\lambda$ denotes the wavenumber and $\lambda$ the wavelength. The formulation extends naturally to three-dimensional settings and other wave phenomena. Wave excitation is imposed through prescribed field values at source-associated locations rather than through an explicit forcing term in Eq.~\ref{eq:helmholtz}. To emulate an unbounded medium, the outer boundary $\Gamma_{\mathrm{ext}}$ of the truncated computational domain satisfies the first-order radial absorbing condition. 

PE-PINN represents the field as a superposition of physics-guided kernel-envelope components, 
\noindent
\begin{minipage}{0.58\linewidth}
\begin{equation}
E_z(\mathbf{x}) = \sum_{m=1}^{M} w_m(\mathbf{x})\,A_m(\mathbf{x})\,\Psi_m(\mathbf{x}),
\label{eq:kernel_envelope}
\end{equation}
\end{minipage}%
\hfill
\begin{minipage}{0.40\linewidth}
\begin{equation}
\Psi_m(\mathbf{x}) = e^{-jk\|\mathbf{x}-\mathbf{x}_m\|_2}.
\label{eq:primitive_kernel}
\end{equation}
\end{minipage}
In Eq.~\ref{eq:kernel_envelope}, $\Psi_m$ is an analytically prescribed propagation kernel, $A_m$ is a learnable envelope that captures the remaining smooth variation, and $w_m\in(0,1)$ is a spatial gating function that restricts each component to its region of validity. For a point-like source located at $\mathbf{x}_m$, the primitive kernel takes the spherical-wave form in Eq.~\ref{eq:primitive_kernel}.
Because the rapid oscillatory phase is encoded directly in $\Psi_m$, the network learns only the smoother envelopes $A_m$, greatly reducing the burden imposed by spectral bias. The effectiveness of the representation, however, depends critically on the completeness of the kernel dictionary. Any physical field component not captured by the kernel set must be reproduced by the envelopes, placing an implicit limit on the attainable accuracy.

We consider wave fields generated by \emph{structured} source systems exhibiting recursive geometric organization, including dipoles, antenna and microphone arrays, phased arrays, metasurfaces, and distributed sensing platforms. Such systems can be constructed hierarchically: a larger assembly is formed from translated copies of a smaller subsystem, ultimately traceable to a single elementary unit. For example, a dipole consists of two point sources, while a dipole array consists of translated copies of dipoles.
Let $\mathcal{C}^{(\ell)}$ denote the source configuration at hierarchy level $\ell$, where $\mathcal{C}^{(0)}$ corresponds to a single elementary unit. The transition from level $\ell$ to $\ell+1$ is described by 
\begin{equation} 
\mathcal{C}^{(\ell+1)} = \bigcup_{i=1}^{I_{\ell}} T_i^{(\ell)} \!\left( \mathcal{C}^{(\ell)} \right), \label{eq:structured_source} 
\end{equation} 
where $I_\ell$ is the branching factor and $T_i^{(\ell)}$ denotes the spatial transformation associated with the $i$-th copy. Although we focus on translations, the formulation extends directly to general rigid transformations. 
Let $N_\ell$ denote the number of elementary units contained in $\mathcal{C}^{(\ell)}$. From Eq.~\ref{eq:structured_source}, $N_{\ell+1} = I_\ell N_\ell$,  which yields 
\begin{equation} 
N_\ell = \prod_{m=0}^{\ell-1} I_m \;\;\xrightarrow[\;I_m\equiv I\;]{} \;\; I^\ell . 
\label{eq:unit_count}
\end{equation} 
Thus, the number of elementary units grows geometrically with the depth of the hierarchy. 

\subsection{The Kernel Scalability Bottleneck}\label{sec:scalability}

As shown earlier in Table~\ref{tab:helmholtz_comparison}, state-of-the-art PINNs that perform well on manufactured Helmholtz benchmarks often fail on practical radiation problems, even for a simple dipole consisting of only two point sources. PE-PINN overcomes this limitation through the kernel--envelope representation of Eq.~\ref{eq:kernel_envelope}. However, extending this representation to larger source systems incurs a cost that scales with the size of the configuration.
For a given source type, assume each elementary unit requires $q$ primitive kernels. A direct PE-PINN representation of $\mathcal{C}^{(\ell)}$ then requires 
\begin{equation}
    K_{\mathrm{direct}}^{(\ell)} = q N_\ell = q\prod_{m=0}^{\ell-1} I_m
\label{eq:direct_count}
\end{equation}
active kernels. Each kernel introduces an additional carrier-envelope branch in Eq.~\ref{eq:kernel_envelope}, together with its associated envelope network, gating function, and reconstruction operations.
Empirically (Sec.~\ref{sec:experiments}), the per-iteration training cost scales approximately linearly with the number of active kernels. Assuming that the number of training epochs remains roughly constant as the hierarchy expands, the total training cost can be approximated by $
\mathcal{T}^{(\ell)} \sim c\,K_{\mathrm{direct}}^{(\ell)} = c\,q\,N_\ell$,
where $c$ is a problem-dependent constant. By Eq.~\ref{eq:unit_count}, the cost is linear in the number of elementary units but exponential in hierarchical depth. In practice, the scaling can be worse, because larger kernel sets increase both the parameter count and the coupling among envelope branches through the shared PDE residual, thereby making optimization increasingly difficult. 

More fundamentally, a direct representation ignores the structure present in the source hierarchy. The $I_\ell$ sub-configurations comprising level $\ell+1$ are translated copies of one another, and the fields they generate are therefore related by known spatial transformations. Nevertheless, a direct construction learns the corresponding envelopes independently. Once the field generated by a lower-level subsystem has been learned, its representation can itself serve as a reusable building block for higher levels. In other words, the hierarchy in the physical source configuration induces a corresponding hierarchy in the solution space, where a level-$\ell$ solution can be abstracted as a single reusable component rather than expanded into $N_\ell$ primitive kernels. This observation motivates the central objective of this work. We seek a representation whose number of trainable kernel branches grows with the \emph{depth} of the hierarchy rather than with the \emph{number} of elementary units it contains; that is, $\mathcal{O}(\ell)$ instead of $\mathcal{O}(I^\ell)$. At the same time, the representation must retain the convergence advantages provided by architectural physics embedding.

\begin{wrapfigure}{r}{0.58\textwidth}
\vspace{-22mm}
    \centering
    \includegraphics[width=1\linewidth]{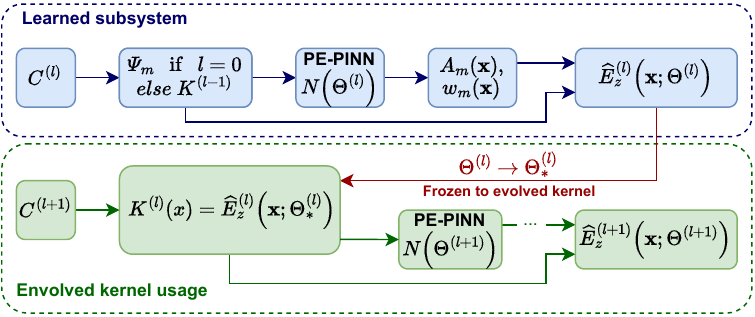}
    \caption{Overview of PE-EK-PINN structure.}
    \label{fig:structure_overview}
\vspace{-2mm}
\end{wrapfigure}

\section{PE-EK-PINN}\label{sec:cascade_method}

\subsection{Overview}

Figure~\ref{fig:structure_overview} illustrates the PE-EK-PINN framework. At hierarchy level $\ell$, the source configuration $\mathcal{C}^{(\ell)}$ is learned using either primitive physics kernels or an evolved kernel inherited from level $\ell-1$. After convergence, the resulting field representation is frozen and abstracted into a reusable kernel $K^{(\ell)}$, which serves as a fixed physics-aware building block for learning the next-level configuration $\mathcal{C}^{(\ell+1)}$. By repeatedly applying this \emph{learn-freeze-reuse} process, PE-EK-PINN constructs a hierarchy of increasingly expressive kernels while preserving all previously learned representations. Consequently, physical structure discovered at one scale is reused at higher levels rather than being repeatedly relearned.

\subsection{Evolving Physics Kernels}

The predicted wave field is represented as a superposition of physics-guided kernel (Eq.~\ref{eq:kernel_envelope}). Primitive kernels are chosen according to the underlying wave physics. All higher-level kernels are evolved recursively from this primitive representation. Consider the subsystem associated with $\mathcal{C}^{(\ell)}$. Training a PE-PINN on this subsystem yields a field approximation $\hat{E}_z^{(\ell)}(\mathbf{x};\Theta^{(\ell)})$, where $\Theta^{(\ell)}$ denotes the trainable parameters at hierarchy level $\ell$. After convergence, the optimized parameters $\Theta_*^{(\ell)}$ are frozen and the learned field is promoted to an \emph{evolved kernel}, 
\begin{equation} 
K^{(\ell)}(\mathbf{x}) = \hat{E}_z^{(\ell)} \!\left( \mathbf{x}; \Theta_*^{(\ell)} \right). 
\label{eq:cached_kernel} 
\end{equation}

Unlike the primitive kernel, $K^{(\ell)}$ is both \emph{composite} and \emph{learned}. It encapsulates the collective wave behavior of the entire subsystem, including all lower-level kernel interactions used to represent $\mathcal{C}^{(\ell)}$, and it derives its expressive power from the converged solution rather than from an analytical formula. Nevertheless, it serves the same functional role in Eq.~\ref{eq:kernel_envelope} by capturing the dominant oscillatory structure of the field so that the network at the next hierarchy level only needs to learn a comparatively smooth envelope. In this way, complex wave behavior discovered at one level becomes a reusable physics-aware building block for the next.

\subsection{Cascaded Kernel Reuse}

Kernel reuse across spatial locations is enabled by the translation invariance of the Helmholtz operator. In a homogeneous medium with constant wavenumber $k$, if $K^{(\ell)}$ satisfies $\nabla^2 K^{(\ell)} + k^2 K^{(\ell)} = 0$, then any translated copy satisfies the same equation. For the $m$-th instance of a level-$\ell$ subsystem, we therefore define the translated evolved kernel 
\begin{equation} 
K_m^{(\ell)}(\mathbf{x}) = K^{(\ell)} \!\left( \mathbf{x}-\Delta\mathbf{c}_m^{(\ell)} \right), 
\label{eq:translated_cached_kernel} 
\end{equation} 
where $\Delta\mathbf{c}_m^{(\ell)}$ is the displacement from the reference subsystem to its $m$-th copy, as specified by the transformation $T_m^{(\ell)}$ in Eq.~\ref{eq:structured_source}. 

A level-$(\ell+1)$ configuration is then represented as a superposition of the $I_\ell$ translated kernels, 
\begin{equation} 
\hat{E}_z^{(\ell+1)}(\mathbf{x}) = \sum_{m=1}^{I_{\ell}} w_m^{(\ell+1)}(\mathbf{x}) A_m^{(\ell+1)}(\mathbf{x}) K_m^{(\ell)}(\mathbf{x}), 
\label{eq:cascade_representation} 
\end{equation} 
where the envelopes $A_m^{(\ell+1)}$ and gates $w_m^{(\ell+1)}$ are learned at the current level, while the evolved kernels remain frozen. The envelopes are still required because a simple superposition of isolated subsystem fields is generally not the solution of the composite problem. Interactions among subsystems, together with the source constraints on individual elements and the absorbing condition on the outer boundary, must still be satisfied. The evolved kernels provide the dominant oscillatory structure, while the envelopes capture these interaction effects. 

After the level-$(\ell+1)$ model converges, its learned field is frozen and promoted to a new evolved kernel, allowing the process to recurse: 
\begin{equation} 
\text{primitive kernels} \rightarrow K^{(0)} \rightarrow K^{(1)} \rightarrow \cdots \rightarrow K^{(L)}. 
\label{eq:cascade_hierarchy} 
\end{equation}
Through this cascaded construction, groups of primitive kernels are progressively compressed into a small number of increasingly expressive evolved kernels, enabling higher-level structures to be represented and reused without repeatedly expanding them into their constituent primitives.

\subsection{Training and Computational Efficiency}

At each cascade level, only the parameters of the current kernel-envelope representation are optimized. All evolved kernels inherited from lower levels are frozen and excluded from gradient updates. The training objective follows the PE-PINN formulation, 
\begin{equation} 
\mathcal{L} = \lambda_{\mathrm{src}}\mathcal{L}_{\mathrm{src}} + \lambda_{\mathrm{pde}}\mathcal{L}_{\mathrm{pde}} + \lambda_{\mathrm{bc}}\mathcal{L}_{\mathrm{bc}}, 
\label{eq:cascade_loss} 
\end{equation} 
where $\mathcal{L}_{\mathrm{src}}$, $\mathcal{L}_{\mathrm{pde}}$, and $\mathcal{L}_{\mathrm{bc}}$ enforce the source excitation, Helmholtz residual, and absorbing boundary condition, respectively. Importantly, $\mathcal{L}_{\mathrm{pde}}$ is evaluated on the composite field of Eq.~\ref{eq:cascade_representation} over the entire level-$(\ell+1)$ domain. Thus, physical consistency is re-enforced at every hierarchy level rather than merely inherited from previously cached kernels.
As established in Sec.~\ref{sec:scalability}, a direct PE-PINN representation of the level-$L$ configuration $\mathcal{C}^{(L)}$ requires $K_{\mathrm{direct}}^{(L)} = qN_L$ simultaneously active kernels, where $N_L$ is the number of elementary units in the configuration.
Under the proposed cascading scheme, level $\ell+1$ is trained using only $I_\ell$ active kernels because each lower-level subsystem is represented by a single frozen evolved kernel. Consequently, the maximum number of active kernels encountered during the entire training curriculum is $
K_{\mathrm{cascade}}^{\mathrm{peak}} = \max\!\left( q,\; \max_{\ell} I_\ell \right)$, 
which is independent of $N_L$. As a result, the peak memory footprint and per-iteration training cost remain bounded even as the physical system grows. 

Since cascading trains $L+1$ models sequentially, the total training effort is better characterized by the cumulative active-kernel count across all levels. Assuming that the number of training epochs per level is approximately constant and that per-iteration cost scales linearly with the number of active kernels, the cumulative cost becomes 
\begin{equation} 
K_{\mathrm{cascade}}^{\mathrm{total}} = q+\sum_{\ell=0}^{L-1} I_\ell \;\;\xrightarrow[\;I_\ell\equiv I\;]{} \;\; q+IL = q+I\log_I N_L ,
\label{eq:cascade_total} 
\end{equation}
where the final equality follows from $N_L=I^L$. The proposed hierarchy therefore reduces the dependence on system size from $\mathcal{O}(N)$ in Eq.~\ref{eq:direct_count} to $\mathcal{O}(\log N)$, converting the exponential growth with hierarchy into linear growth. Experimental results in Sec.~\ref{sec:experiments} validate this scaling behavior.

\section{Experiments and Evaluation}\label{sec:experiments}

\subsection{Experiment Setup}

We construct all configurations from two elementary two-dimensional radiators: a scalar dipole and a uniform finite line source. Both admit closed-form outgoing Helmholtz solutions in terms of Hankel functions of the second kind, with the dipole derived from $H_1^{(2)}$ and the line source obtained by integrating the Green's function $\tfrac{\mathrm{j}}{4}H_0^{(2)}$ along the segment. These analytical fields define both the source conditions during training and the evaluation references, eliminating numerical-solver and discretization errors. Because both fields are singular on the source support, excitation is imposed on enclosing boundaries: a circle of radius $0.5\lambda$ for the dipole and a capsule of radius $0.05\lambda$ for the line source. PDE collocation points are sampled outside slightly larger exclusion regions. Each field is normalized to unit magnitude on its source boundary, and multi-element configurations are formed by coherent superposition under uniform unit excitation. The resulting superposed field serves as both the source boundary target and the evaluation reference. For PE-PINN, a $2\lambda$ line source uses ten spherical kernels and a $0.5\lambda$ line source uses four. These kernels are representation components in Eq.~\ref{eq:kernel_envelope}, not discrete physical radiators. Complete derivations, normalization constants, exclusion radii, and boundary specifications are provided in Appendix~\ref{app:source_reference}.

\begin{table}[t]

\centering
\caption{Overview of the evaluated source configurations.}
\label{tab:scenario_overview}
\small
\setlength{\tabcolsep}{4pt}
\begin{tabular}{l l l l}
\toprule
\textbf{Scenario} & \textbf{Base config.} & \textbf{PE-EK-PINN construction} & \textbf{Transformation} \\
\midrule

Dipole array & Single dipole & $1 \rightarrow 2{\times}2 \rightarrow 4{\times}4 \rightarrow 8{\times}8 \rightarrow 16{\times}16$ & Translation \\

Composite line geometry & $2\lambda$ line & $2\lambda$ line $\rightarrow$ \{Cross, 5-point star\} & Translation + rotation \\

Cross array (Section~\ref{app:cross_array}) & $0.5\lambda$ line & $0.5\lambda$ line $\rightarrow$ Cross $\rightarrow 2{\times}2 \rightarrow 4{\times}4$ & Translation + rotation \\
\bottomrule

\vspace{-8mm}

\end{tabular}
\end{table}

Table~\ref{tab:scenario_overview} lists the three families of structured configurations evaluated: dipole arrays, composite finite-line geometries, and cross arrays built from shorter line sources. Each admits a consistent analytical reference at every level of its hierarchy, so the predicted complex field can be compared against the exact solution at all cascade stages.
All experiments are conducted in free space at $f=2.4$~GHz over the domain $[-2.5,2.5]^2~\mathrm{m}^2$. PDE collocation points are defined on a uniform $0.01$~m grid (approximately $12.5$ points per wavelength), excluding points within the source exclusion regions. The source constraints are imposed on all sampled source-boundary points; hence, $N_{\mathrm{src}}$ varies with the source geometry and array size. Each of the four outer boundaries is also uniformly sampled with a spacing of $0.01$~m, resulting in $501$ points per edge and $N_{\mathrm{bc}}=2{,}004$ boundary samples in total. All configurations use the same PE-PINN backbone with hidden widths $[40,120,120,120]$ and sinusoidal activations. Each active kernel is associated with an independent output head that predicts the real and imaginary parts of its envelope together with a spatial gating term. Models are trained using Adam with a learning rate of $10^{-4}$ for $50{,}000$ iterations. Training minimizes the objective in Eq.~\ref{eq:cascade_loss}. The source loss enforces complex-field matching on the source boundary, with an additional normal-derivative constraint for the finite line source. The Helmholtz residual is evaluated outside the source-exclusion regions, and a first-order radial outgoing condition is imposed on the outer boundary. 
The loss in Eq.~\ref{eq:cascade_loss} uses fixed weights $\lambda_{\mathrm{pde}}=0.01$, $\lambda_{\mathrm{src}}=10$, and $\lambda_{\mathrm{bc}}=1$, chosen to balance the numerical scales of the three residual terms. The complete loss definitions are provided in Appendix~\ref{app:training_constraints}. These hyperparameters are kept unchanged across all configurations and cascade levels. All experiments are conducted on a workstation equipped with an NVIDIA GeForce RTX 4090 GPU and a 13th Gen Intel Core i9-13900K CPU.

\subsection{Evaluation Results}


We compare PE-EK-PINN with PE-PINN using primitive kernels under identical settings. Since coherent superposition increases field magnitude with configuration size, we use complex relative $L_2$ error as the primary metric and report MSE as a secondary reference. For PE-EK-PINN, stage time denotes the current cascade level, while cumulative time includes all preceding levels.

\subsubsection{Dipole Array}
\label{sec:dipole_results}

Dipoles are identically oriented on a square lattice with $1\lambda$ center-to-center spacing. Four translated copies of the learned single-dipole kernel form the $2\times2$ array; the converged $2\times2$ model is then frozen and reused for the $4\times4$, and so on. Every cascade level therefore carries exactly four active kernels, independent of the $4$ to $256$ physical dipoles in the target array.

\begin{table}[ht!]

\centering
\caption{Accuracy and training time for the dipole-array experiments. PE-PINN results for the
$2\times2$ and $4\times4$ configurations serve as an ablation baseline. Cumulative time includes all
preceding cascade stages starting from the single-dipole model.}
\label{tab:dipole_results}
\begin{tabular}{lcccccc}
\toprule
Method & $N_{\mathrm{dipoles}}$ & $N_{\mathrm{kernels}}$ & Stage T. & Cum. T. & MSE & Rel. $L_2$ \\
\midrule
PE-PINN    & $1$          & 2  & 00:19:47 & N/A & $9.48\times10^{-6}$ & $2.14\times10^{-2}$ \\
PE-EK-PINN & $2\times2$   & 4  & 00:26:51 & 00:46:38 & $2.49\times10^{-5}$ & $1.55\times10^{-2}$ \\
PE-EK-PINN & $4\times4$   & 4  & 00:26:53 & 01:13:31 & $1.86\times10^{-4}$ & $1.62\times10^{-2}$ \\
PE-EK-PINN & $8\times8$   & 4  & 00:27:29 & 01:41:00 & $3.78\times10^{-3}$ & $2.71\times10^{-2}$ \\
PE-EK-PINN & $16\times16$ & 4  & 00:29:35 & 02:10:35 & $3.20\times10^{-1}$ & $9.43\times10^{-2}$ \\
\midrule
PE-PINN    & $2\times2$   & 8  & 01:07:48 & N/A & $5.83\times10^{-5}$ & $2.37\times10^{-2}$ \\
PE-PINN    & $4\times4$   & 32 & 04:27:47 & N/A & $3.49\times10^{-4}$ & $2.22\times10^{-2}$ \\
\bottomrule

\vspace{-8mm}

\end{tabular}
\end{table}

\begin{figure}[b]

    \centering
    \includegraphics[width=.99\linewidth]{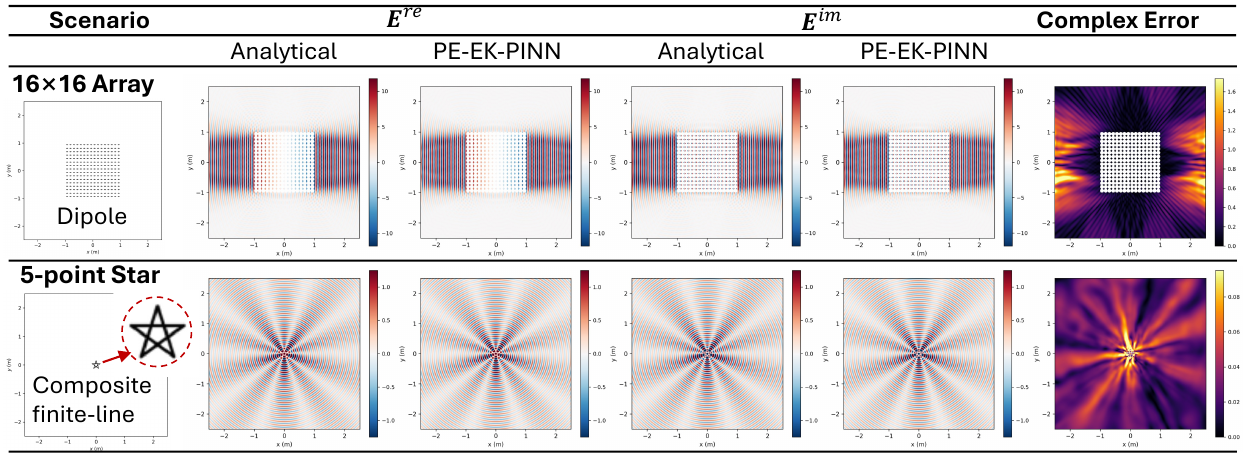}
    \caption{Visualized results of the $16\times 16$ dipole-array and 5-point star finite-line experiments. Please refer to Appendix~\ref{sec:app:visual} for the complete visualized results.}
    \label{fig:visual_main}
\end{figure}

As shown in Table~\ref{tab:dipole_results}, the relative $L_2$ error remains below $3\%$ through the $8\times8$ array ($1.55\times10^{-2}$, $1.62\times10^{-2}$, and $2.71\times10^{-2}$), increasing to $9.43\times10^{-2}$ only for the $16\times16$ case. This degradation is likely due to two factors: the accumulation of approximation error across multiple frozen hierarchy levels and the increasing mismatch between a spatially extended array and the single-center radial absorbing boundary condition, which assumes that outgoing waves originate from a single location. 
PE-EK-PINN is not only more efficient than PE-PINN but also more accurate (in terms of relative $L_2$ errors) on the same configurations. Figure~\ref{fig:visual_main} shows the visualized results for the $16\times16$ case. The likely reason for the higher accuracy is that each evolved kernel already captures the converged field of a subsystem, allowing the next level to optimize only a small number of smooth envelopes rather than many coupled kernel-envelope branches.

Training cost follows the scaling analysis of Sec.~\ref{sec:cascade_method}. Increasing the number of active kernels in PE-PINN from $8$ to $32$ increases training time by $3.98\times$, confirming the near-linear dependence on active-kernel count. In contrast, PE-EK-PINN exhibits nearly constant stage time, ranging from $26.9$ to $29.6$ minutes from the $2\times2$ to the $16\times16$ array despite a $64\times$ increase in the number of physical dipoles. A similar scaling trend is observed for the hierarchical cross-array experiment (Appendix~\ref{app:cross_array}). 


\subsubsection{Composite Finite-Line Geometries}
\label{sec:line_results}


The second experiment reuses a pretrained $2\lambda$ continuous line source under rotation and translation. The line source uses ten primitive kernels and requires 1:37:08 of training before being frozen as an evolved kernel. Table~\ref{tab:line_results} summarizes the results on the evaluated composite finite-line geometries described below.

\textbf{Two-line cross.} A cross is formed from two perpendicular copies of the pretrained line kernel and compared against a PE-PINN baseline using $20$ primitive point-source kernels. PE-EK-PINN requires only two active kernels, reducing stage training time from $3{:}13{:}19$ to $0{:}16{:}29$ ($11.73\times$ faster), or $1.70\times$ end-to-end when the $1{:}37{:}08$ base-kernel training cost is included. The relative $L_2$ error decreases from $1.33\times10^{-1}$ to $3.34\times10^{-2}$, the largest accuracy improvement among the evaluated configurations. The PE-PINN baseline again validates the cost model: doubling the active kernels from $10$ to $20$ increases training time by $1.99\times$.

\textbf{5-point star and kernel validity.} The star is constructed from five translated line kernels, each rotated into the reference frame of the pretrained horizontal kernel. Rotation expands the effective coordinate range beyond the original training domain $[-2.5,2.5]^2$, requiring kernel evaluations at coordinates approaching $\pm3.53$. Since the kernel is represented by an MLP, such evaluations constitute extrapolation and are unreliable. To avoid this issue, we pretrain a separate line kernel on the enlarged domain $[-3.6,3.6]^2$.
The expanded domain spans $57.6$ rather than $40$ wavelengths. Although the kernel count remains ten, training time increases by $1.46\times$ ($2{:}21{:}57$ versus $1{:}37{:}08$), and the base relative $L_2$ error rises from $4.12\times10^{-2}$ to $7.56\times10^{-2}$.  Figure~\ref{fig:visual_main} shows the visualized results for the 5-point star case. Starting from this base, the star requires only $0{:}39{:}18$ of stage training with five active kernels and achieves a relative $L_2$ error of $6.32\times10^{-2}$, for a cumulative training time of $3{:}01{:}14$. No PE-PINN baseline is reported because a direct representation would require $50$ primitive kernels, which is beyond a practical training budget.

\begin{table}[t]
\centering

\caption{Accuracy and training time for the finite-line-source experiments. The wider-domain line
kernel is trained on $\Omega=[-3.6,3.6]^2$ to cover the coordinate range reached after rotation.}
\label{tab:line_results}
\begin{tabular}{llccccc}
\toprule
Method & Config. & Kernels & Stage T. & Cum. T. & MSE & Rel. $L_2$ \\
\midrule
PE-PINN    & Line              & 10 & 01:37:08 & N/A & $8.85\times10^{-5}$ & $4.12\times10^{-2}$ \\
PE-EK-PINN & Two-line cross             & 2  & 00:16:29 & 01:53:37 & $1.18\times10^{-4}$ & $3.34\times10^{-2}$ \\
PE-PINN    & Two-line cross             & 20 & 03:13:19 & N/A      & $1.87\times10^{-3}$ & $1.33\times10^{-1}$ \\
PE-PINN    & Line$_{\rm wide}$ & 10 & 02:21:57 & N/A & $2.07\times10^{-4}$ & $7.56\times10^{-2}$ \\
PE-EK-PINN & 5-point star      & 5  & 00:39:18 & 03:01:14 & $1.24\times10^{-3}$ & $6.32\times10^{-2}$ \\
\bottomrule

\vspace{-8mm}

\end{tabular}
\end{table}

\subsection{Discussion}
PE-EK-PINN is based on a simple but effective idea that a converged subsystem field can itself serve as a reusable physics kernel. The computational gain comes from replacing many trainable kernel-associated envelope branches and their gradients with a single fixed component. Consequently, the observed speedup scales more slowly than the reduction in kernel count, but improves with hierarchy depth as increasingly large subsystems are absorbed into each cached kernel. 

Because an evolved kernel is only an approximation, its error propagates to higher levels. Later envelopes can compensate for interactions among reused components but cannot correct inaccuracies internal to a frozen kernel. Error accumulation is modest for shallow hierarchies but becomes more noticeable at greater depths, making the accuracy of the base kernel particularly important.

Kernel reuse also relies on the translation invariance of the Helmholtz operator in homogeneous media. In addition, since a cached kernel is represented by an MLP, it is reliable only within the coordinate range covered during training. Translations or rotations may map evaluation points beyond this region and lead to extrapolation errors, as demonstrated in Sec.~\ref{sec:line_results}. Expanding the base training domain can mitigate this issue, but increases training cost and problem difficulty.

Finally, PE-EK-PINN assumes a known hierarchy and therefore provides limited compression for configurations without repeated substructure. Automatically discovering reusable hierarchies remains an important direction for future work. PE-EK-PINN is also complementary to existing scalable PINN techniques. For example, domain decomposition exploits spatial locality, separable architectures exploit low-rank solution structure, while PE-EK-PINN exploits redundancy in the source configuration.

\section{Conclusion}

We proposed PE-EK-PINN, a hierarchical framework that promotes learned subsystem fields into reusable physics kernels. By reusing pretrained subsystem representations, PE-EK-PINN keeps the number of active kernels bounded and reduces cumulative training growth from $\mathcal{O}(N)$ to $\mathcal{O}(\log N)$. Experiments on dipole arrays, cross arrays, and composite line geometries validate this scaling while improving both training efficiency and reconstruction accuracy over direct PE-PINN. These results demonstrate the potential of reusable learned kernels for scaling physics-informed wave modeling to larger structured systems.


\textbf{AI Use Disclosure:} In this work, we have not used generative AI tools for tasks that require disclosure. Additionally, we used generative AI tools to assist with creating, editing, and debugging software code, as well as with editing a research paper to improve readability. All AI-assisted code was reviewed, verified, and tested for correctness by the authors. We take responsibility for the final content of this work, including text, claims, or artifacts produced with the aid of generative AI.

\subsubsection*{Acknowledgments}
TBA

\bibliography{iclr2027_conference}
\bibliographystyle{iclr2026_conference}

\appendix
\section{Source Definitions and Analytical Reference Solutions}
\label{app:source_reference}

This appendix gives the closed-form fields, normalization conventions, and boundary constraints. Throughout, $\lambda$ denotes the wavelength,
$k=2\pi/\lambda$, and $H_n^{(2)}$ the $n$-th order Hankel function of the second kind, whose asymptotic
behavior corresponds to an outgoing wave under the $e^{\mathrm{j}\omega t}$ time convention.

\subsection{Dipole Source}
\label{app:dipole}

For a dipole centered at $\mathbf{x}_s=(x_s,y_s)$ with unit orientation
$\mathbf{p}=(\cos\alpha,\sin\alpha)$, the outgoing analytical field is
\begin{equation}
E_z^{\mathrm{dipole}}(x,y) = -\frac{\mathrm{j}k}{4}\, H_1^{(2)}(kr)\,
\frac{\cos\alpha\,(x-x_s) + \sin\alpha\,(y-y_s)}{r},
\qquad
r = \sqrt{(x-x_s)^2 + (y-y_s)^2},
\label{eq:analytic_dipole}
\end{equation}
in which the trailing factor is the projection $\mathbf{p}\cdot(\mathbf{x}-\mathbf{x}_s)/r$ giving the
$\cos$ radiation pattern.

\paragraph{Singularity handling.}
Equation~\ref{eq:analytic_dipole} is singular at $\mathbf{x}_s$, so the source condition is not
imposed at the dipole center. Instead, analytical field values are prescribed on a circular source
boundary of radius
\begin{equation}
r_{\mathrm{src}}^{\mathrm{dipole}} = 0.5\lambda ,
\end{equation}
and the homogeneous Helmholtz residual is enforced only outside the slightly larger exclusion region
\begin{equation}
r_{\mathrm{exc}}^{\mathrm{dipole}} = 0.51\lambda ,
\label{eq:dipole_exclusion_radius}
\end{equation}
so that no PDE collocation point falls inside the singular source region.

\paragraph{Normalization.}
The field is scaled by its peak radial magnitude on the source boundary, obtained by setting the
angular factor in Eq.~\ref{eq:analytic_dipole} to unity:
\begin{equation}
C_{\mathrm{dipole}} = \left| -\frac{\mathrm{j}k}{4}\,
H_1^{(2)}\!\left( k r_{\mathrm{src}}^{\mathrm{dipole}} \right) \right| ,
\qquad
\widetilde{E}_z^{\mathrm{dipole}}(x,y) = \frac{E_z^{\mathrm{dipole}}(x,y)}{C_{\mathrm{dipole}}} .
\label{eq:dipole_normalization}
\end{equation}

\paragraph{Dipole arrays.}
For a configuration of $N_d$ dipoles, the analytical total field follows by coherent superposition of
the normalized element fields,
\begin{equation}
E_z^{\mathrm{dipole\text{-}array}}(x,y) = \sum_{n=1}^{N_d} a_n\,
\widetilde{E}_{z,n}^{\mathrm{dipole}}(x,y),
\label{eq:dipole_array_field}
\end{equation}
where $a_n\in\mathbb{C}$ is the prescribed complex excitation of the $n$-th element. All dipole-array
experiments in this work use identical excitations, $a_n = 1+0\mathrm{j}$ for $n=1,\ldots,N_d$.
Equation~\ref{eq:dipole_array_field} is used both to define the source targets on every dipole
source boundary and as the analytical reference for evaluation.

\subsection{Uniform Finite Line Source}
\label{app:line}

We consider a uniform continuous line source of length $L$, centered at $(x_c,y_c)$ and oriented at
angle $\alpha$. The analytical reference field is obtained by integrating the two-dimensional outgoing
Green's function along the segment,
\begin{equation}
E_z^{\mathrm{line}}(x,y) = q \int_{-L/2}^{L/2} \frac{\mathrm{j}}{4}\,
H_0^{(2)}\!\left( kR(s) \right) \,\mathrm{d}s ,
\label{eq:analytic_line_field}
\end{equation}
where $q\in\mathbb{C}$ is the uniform line-source density and
\begin{equation}
R(s) = \sqrt{ \left( x-x_c-s\cos\alpha \right)^2 + \left( y-y_c-s\sin\alpha \right)^2 }
\label{eq:line_source_distance}
\end{equation}
is the distance from the field point $(x,y)$ to the source point indexed by arclength $s$. The
integral is evaluated numerically.

\paragraph{Singularity handling.}
Eq.~\ref{eq:analytic_line_field} is singular on the physical source segment, so the source is
enclosed by a thin capsule-shaped region of radius $r_{\mathrm{cap}} = 0.05\lambda$ about the segment.
The reference field is evaluated on the capsule boundary and used there to prescribe the source
condition; PDE collocation points are excluded from the capsule interior.

\paragraph{Normalization.}
The line-source field is globally normalized so that its RMS magnitude over the $N_{\mathrm{src}}$
source points $\{(x_i,y_i)\}$ sampled on the capsule boundary is unity:
\begin{equation}
\widetilde{E}_z^{\mathrm{line}}(x,y) = \frac{E_z^{\mathrm{line}}(x,y)}
{\left( \dfrac{1}{N_{\mathrm{src}}} \displaystyle\sum_{i=1}^{N_{\mathrm{src}}}
\left| E_z^{\mathrm{line}}(x_i,y_i) \right|^2 \right)^{1/2}} .
\label{eq:normalized_line_field}
\end{equation}
An RMS convention is used here rather than the peak convention of
Eq.~\ref{eq:dipole_normalization} because the magnitude on the capsule boundary is not constant
along the segment.

\paragraph{Boundary constraints.}
Both the analytical field and its outward normal derivative on the capsule boundary are used to define
the source constraints during training. The normal derivative is
\begin{equation}
\frac{\partial E_z^{\mathrm{line}}}{\partial n}
= q \int_{-L/2}^{L/2} -\frac{\mathrm{j}k}{4}\, H_1^{(2)}\!\left(kR(s)\right)
\frac{\left(x-x_c-s\cos\alpha\right)n_x + \left( y-y_c-s\sin\alpha \right)n_y}{R(s)} \,\mathrm{d}s ,
\label{eq:analytic_line_normal_derivative}
\end{equation}
where $(n_x,n_y)$ is the outward unit normal of the capsule boundary, and the same normalization
constant as in Eq.~\ref{eq:normalized_line_field} is applied.

\paragraph{Kernel assignment.}
Two line lengths are used in the experiments. A $2\lambda$ line source is represented by ten spherical wave kernels and a $0.5\lambda$ line source by four, placed along the segment. These kernels are components of the PE-PINN representation in Eq.~\ref{eq:kernel_envelope} rather than discrete physical sources: in both cases the physical source and the reference solution remain the continuous finite line of Eq.~\ref{eq:analytic_line_field}.

\subsection{Training Constraints}
\label{app:training_constraints}

For completeness, we provide the loss terms used in the experiments.

The source loss enforces agreement with the analytical field on the source boundary, 
\begin{equation} 
\mathcal{L}_{E} = \frac{1}{N_{\mathrm{src}}} \sum_{i=1}^{N_{\mathrm{src}}} \left| \hat{E}_z(\mathbf{x}_i)-E_z^{\mathrm{ref}}(\mathbf{x}_i) \right|^2 . 
\label{eq:source_field_loss} 
\end{equation} 
For the dipole, this field-matching term alone defines the source constraint. For the finite line source, whose boundary lies closer to the singular support, the outward normal derivative is also enforced, 
\begin{equation} 
\mathcal{L}_{\mathrm{src}} = \mathcal{L}_{E} + 0.05\,\mathcal{L}_{\partial_n E}, \qquad \mathcal{L}_{\partial_n E} = \frac{1}{k^2N_{\mathrm{src}}} \sum_{i=1}^{N_{\mathrm{src}}} \left| \partial_n\hat{E}_z(\mathbf{x}_i) - \partial_nE_z^{\mathrm{ref}}(\mathbf{x}_i) \right|^2 , 
\label{eq:line_source_loss} 
\end{equation} 
where the factor $k^{-2}$ scales the derivative term to the same order of magnitude as Eq.~\ref{eq:source_field_loss}. The Helmholtz residual is enforced at collocation points outside the exclusion regions, 
\begin{equation} 
\mathcal{L}_{\mathrm{pde}} = \frac{1}{N_{\mathrm{pde}}} \sum_{i=1}^{N_{\mathrm{pde}}} \left| \nabla^2\hat{E}_z(\mathbf{x}_i) + k^2\hat{E}_z(\mathbf{x}_i) \right|^2 . 
\label{eq:pde_loss} 
\end{equation} 
Because the sources are compact and centrally located within a square domain, waves reach the outer boundary predominantly along radial directions rather than the local boundary normal. A conventional normal-derivative absorbing boundary condition therefore becomes less accurate near the corners. To mitigate this effect, we impose a first-order \emph{radial} outgoing condition on all outer boundaries, 
\begin{equation} 
\mathcal{L}_{\mathrm{bc}} = \frac{1}{N_{\mathrm{bc}}} \sum_{i=1}^{N_{\mathrm{bc}}} \left| \frac{\partial\hat{E}_z}{\partial r}(\mathbf{x}_i) + \mathrm{j}k\hat{E}_z(\mathbf{x}_i) \right|^2 , 
\label{eq:bc_loss} 
\end{equation} 
where $\partial/\partial r$ denotes differentiation along the ray extending from the source-configuration center to the boundary point $\mathbf{x}_i$. This radial formulation better matches the dominant propagation direction and reduces artificial reflections from the truncated boundary.

\section{Evaluation Results}\label{sec:app:evaluation}

\subsection{Baselines}

PE-PINN~\cite{zhang2026physicsinformedneuralnetworksarchitectural} is compared with the following works.

\textbf{PINN} \citep{raissi2019pinn} represents the first systematic attempt to integrate governing PDEs directly into neural-network optimization, establishing a foundational framework for physics-informed artificial intelligence. By embedding PDE residuals together with boundary and initial conditions into the training objective, PINN learns continuous solution fields from collocation points without requiring labeled solution data or an explicit computational mesh.

\textbf{gPINN} \citep{yu2022gradient} augments the standard PDE residual loss with derivatives of the residual with respect to the inputs, providing additional differential constraints to improve solution accuracy. 

\textbf{SPINN} \citep{cho2023spinn} factorizes multidimensional inputs along individual coordinate axes and combines the resulting one-dimensional subnetworks through a separable representation, substantially reducing the number of network evaluations required for high-dimensional PDEs; its modified-MLP variant, \textbf{SPINN (m)}, retains the same separable formulation while replacing the standard MLP backbone with a modified MLP for improved approximation accuracy. 

\textbf{AHD-PINN} \citep{dashtbayaz2024physics} studies the residual-loss landscape theoretically, showing that sufficiently wide PINNs can globally minimize the residual under appropriate conditions and establishing activation-function design criteria based on bijective higher-order derivatives for $k$-th-order differential operators. 

\textbf{RoPINN} \citep{wu2024ropinn} extends conventional point-wise optimization from isolated collocation points to their continuous neighborhood regions, improving generalization to the underlying continuous PDE domain, particularly for hidden higher-order constraints. 

\textbf{FPINN} \citep{wu2025deep} incorporates fuzzy neural-network components into PINNs to improve robustness to ambiguous or inaccurate data. 

\textbf{CoPINN} \citep{duan2025copinn} addresses the Unbalanced Prediction Problem by dynamically estimating sample difficulty from gradients of the PDE residual and employing a cognitive training scheduler that progressively optimizes the domain from easy to difficult regions.

\subsection{Cross Array}
\label{app:cross_array}

We also evaluate the proposed hierarchy on cross-array configurations. The cross hierarchy begins from a $0.5\lambda$ continuous line source represented by four primitive point-source kernels. Two perpendicular copies of the learned line kernel form a single cross; the cross is frozen and reused for a $2\times2$ array, which in turn is reused for a $4\times4$ array, with $1\lambda$ spacing between neighboring crosses. This hierarchy exercises rotation in addition to translation.

\begin{table}[htbp!]
\centering
\caption{Accuracy and training time for the hierarchical cross-array experiments. PE-PINN baselines use
primitive point-source kernels only.}
\label{tab:cross_array_results}
\begin{tabular}{llccccc}
\toprule
Method & Config. & Kernels & Stage T. & Cum. T. & MSE & Rel. $L_2$ \\
\midrule
PE-PINN    & $0.5\lambda$ line & 4  & 00:42:59 & 00:42:59 & $6.38\times10^{-5}$ & $6.39\times10^{-2}$ \\
PE-EK-PINN & Cross             & 2  & 00:16:26 & 00:59:25 & $1.07\times10^{-4}$ & $4.18\times10^{-2}$ \\
PE-PINN    & Cross             & 8  & 01:00:50 & N/A      & $5.30\times10^{-4}$ & $9.33\times10^{-2}$ \\
PE-EK-PINN & $2\times2$        & 4  & 00:31:44 & 01:31:09 & $8.00\times10^{-4}$ & $5.07\times10^{-2}$ \\
PE-PINN    & $2\times2$        & 32 & 04:05:07 & N/A      & $3.20\times10^{-3}$ & $1.01\times10^{-1}$ \\
PE-EK-PINN & $4\times4$        & 4  & 00:32:24 & 02:03:33 & $7.57\times10^{-3}$ & $5.90\times10^{-2}$ \\
\bottomrule
\end{tabular}
\end{table}

Table~\ref{tab:cross_array_results} exhibits the same scaling behavior on a more complex hierarchy. Once kernel reuse begins, the stage time remains nearly constant at approximately $32$ minutes for both the $2\times2$ and $4\times4$ arrays, despite a fourfold increase in the number of physical crosses. Relative to PE-PINN, PE-EK-PINN achieves a $3.70\times$ speedup for a single cross and a $7.72\times$ speedup at the $2\times2$ stage ($2.69\times$ end-to-end when all preceding levels are included). At the same time, it substantially improves accuracy, reducing the relative $L_2$ error from $9.33\times10^{-2}$ to $4.18\times10^{-2}$ for a single cross and from $1.01\times10^{-1}$ to $5.07\times10^{-2}$ at the $2\times2$ level. 

Two differences from the dipole hierarchy are noteworthy. First, the relative $L_2$ error remains nearly constant across levels, increasing only from $4.18\times10^{-2}$ to $5.90\times10^{-2}$, and the first cascade level even improves upon its own base model ($6.39\times10^{-2}$). Unlike the dipole case, this hierarchy is shallower and grows to only $16$ physical elements rather than $256$, reducing the accumulation of approximation error across frozen levels. Second, the base level accounts for a large fraction of the total training cost, consuming $43$ of the $124$ cumulative minutes. 

\subsection{Visualization Results}\label{sec:app:visual}

\begin{figure}[ht!]
    \centering
    \includegraphics[width=.95\linewidth]{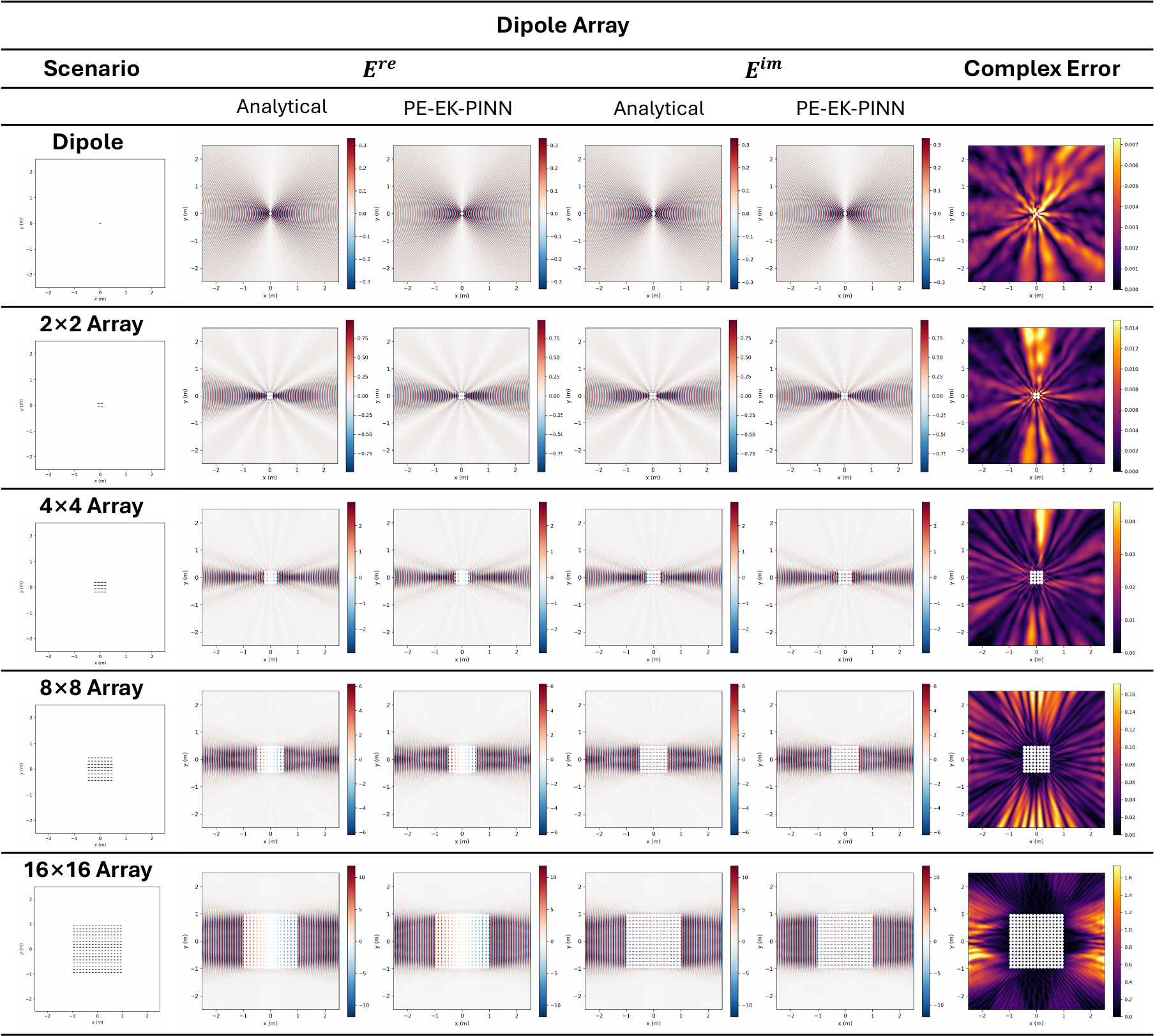}
    \caption{Visualized results of the dipole-array experiments.}
    \label{fig:dipole_array}
\end{figure}

Figure~\ref{fig:dipole_array} illustrates the wave-field evolution across the dipole-array hierarchy. The analytical and PE-EK-PINN fields exhibit consistent spatial and oscillatory structures in both the real and imaginary components as the array expands from a single dipole to the $16\times16$ configuration. The corresponding complex-error maps provide a qualitative view of where the reconstruction differences.

\begin{figure}[htbp!]
    \centering
    \includegraphics[width=.75\linewidth]{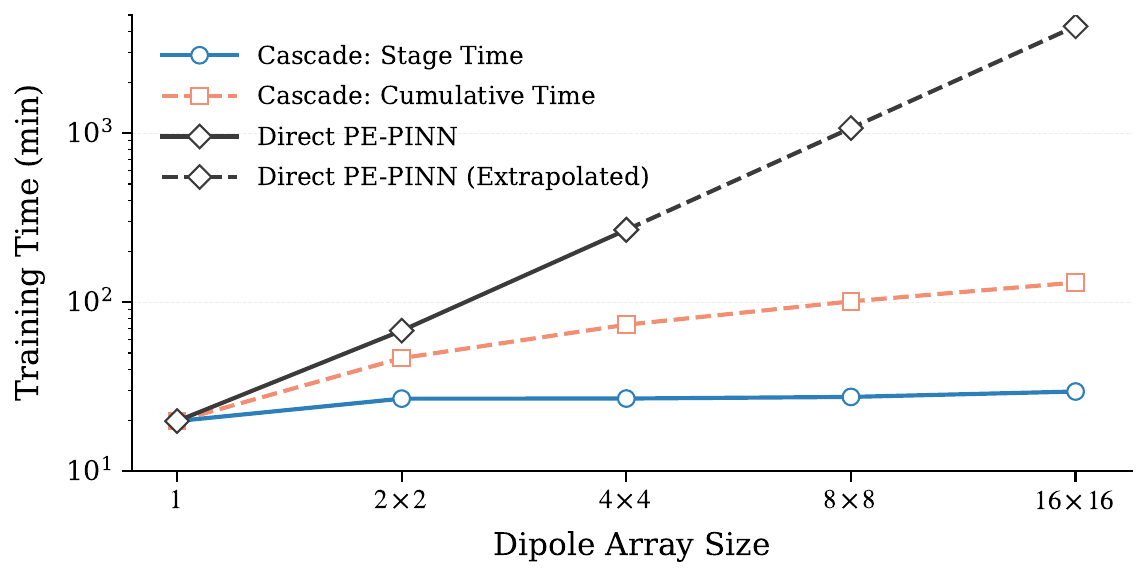}
    \caption{Training-time scaling for direct PE-PINN and PE-EK-PINN. Stage time denotes the training cost of the current cascade level, while cumulative time includes all preceding cascade stages. Direct PE-PINN times for the $8\times8$ and $16\times16$ arrays are extrapolated based on the observed near-linear scaling with active-kernel count.}
    \label{fig:dipole_training_time}
\end{figure}

The direct baseline also reveals an approximately linear dependence of training time on the number of active primitive kernels. Increasing the number of active kernels from 8 to 32 increases the training time by approximately $3.95\times$.  Based on this observed scaling, the direct PE-PINN training times for the $8\times8$ and $16\times16$ arrays are extrapolated to approximately 17.9 and 71.4 hours, respectively. In contrast, the training time of each PE-EK-PINN level remains nearly constant: it varies only from $26.84$ minutes for the $2\times2$ array to $29.59$ minutes for the $16\times16$ array, despite the number of physical dipoles increasing from 4 to 256. Figure~\ref{fig:dipole_training_time} further shows that the cumulative cascade time grows much more slowly than direct training.

\begin{figure}[htbp!]
    \centering
    \includegraphics[width=0.95\linewidth]{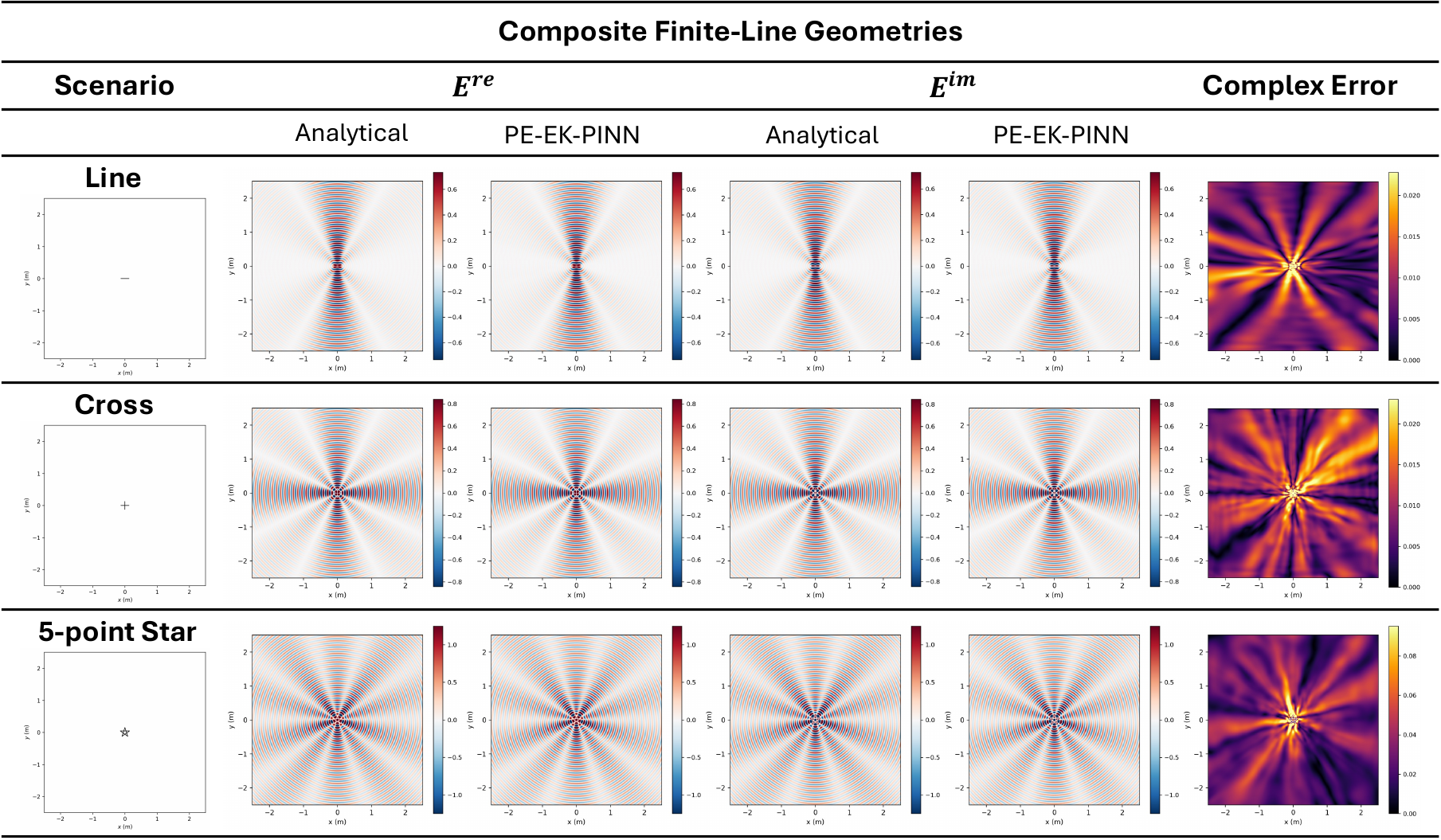}
    \caption{Visualized results of the composite line-source experiments.}
    \label{fig:line_series}
\end{figure}

Figure~\ref{fig:line_series} visualizes the finite-line-source experiments under progressively more complex geometric compositions. A pretrained $2\lambda$ line-source representation is reused through spatial translation and rotation to construct the two-line cross and pentagram configurations. For each geometry, the analytical and PE-EK-PINN solutions show similar spatial wave structures in both the real and imaginary components, illustrating the transfer of the learned line-source kernel to different source arrangements. The complex-error maps provide a qualitative view of the spatial reconstruction differences for the corresponding configurations.

\begin{figure}[ht!]
    \centering
    \includegraphics[width=.95\linewidth]{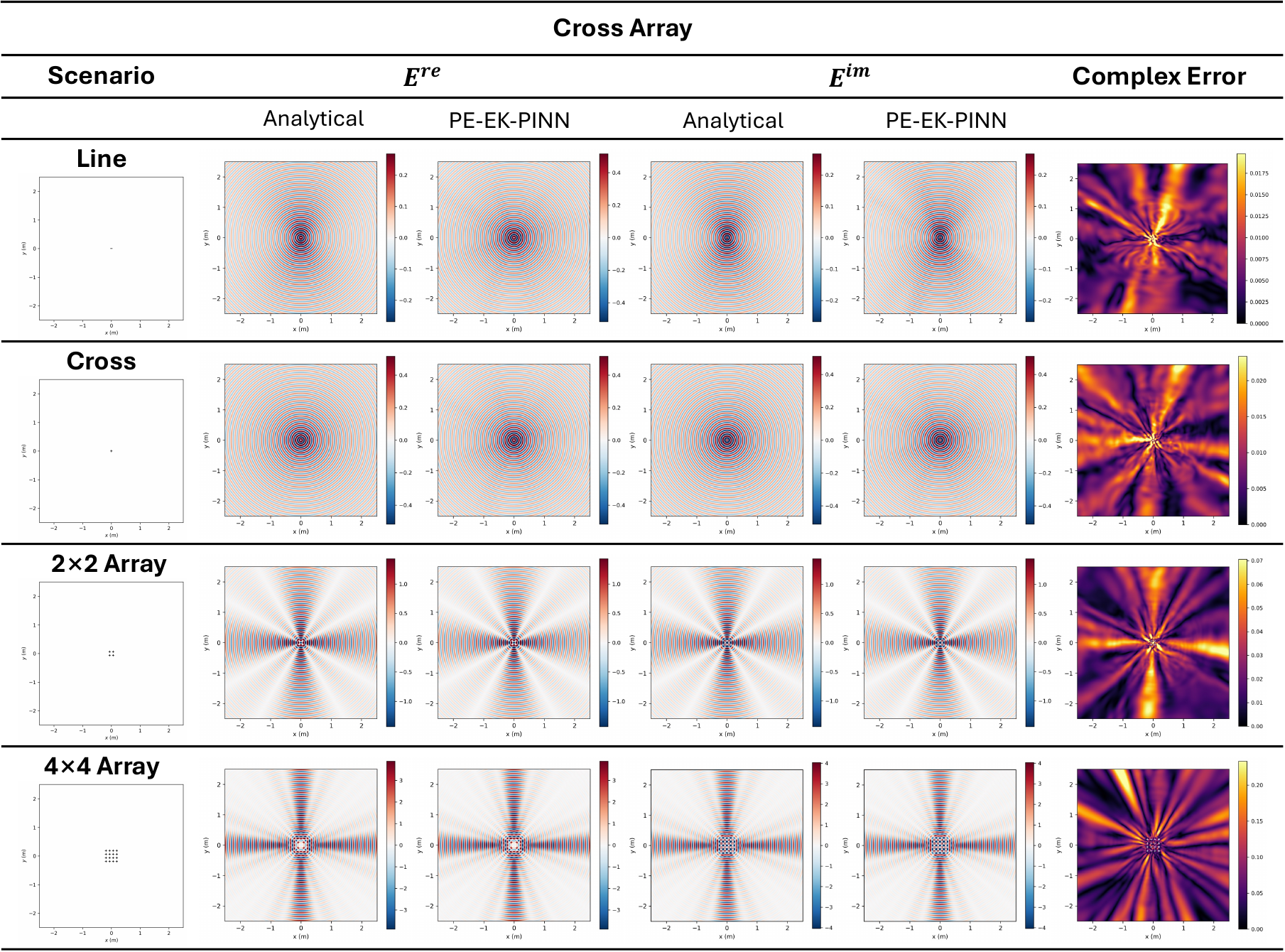}
    \caption{Visualized results the cross-array experiments.}
    \label{fig:cross_array}
\end{figure}

Figure~\ref{fig:cross_array} illustrates the evolution across the cross-array configurations. Starting from the $0.5\lambda$ line source, the learned representation is successively reused to construct a single cross, a $2\times2$ cross array, and a $4\times4$ cross array. The analytical and PE-EK-PINN solutions exhibit consistent spatial interference patterns and oscillatory structures in both the real and imaginary field components across the hierarchy. The corresponding complex-error maps visualize the spatial distribution of the reconstruction differences at each level.

\begin{figure}[ht!]
    \centering
    \includegraphics[width=0.75\linewidth]
    {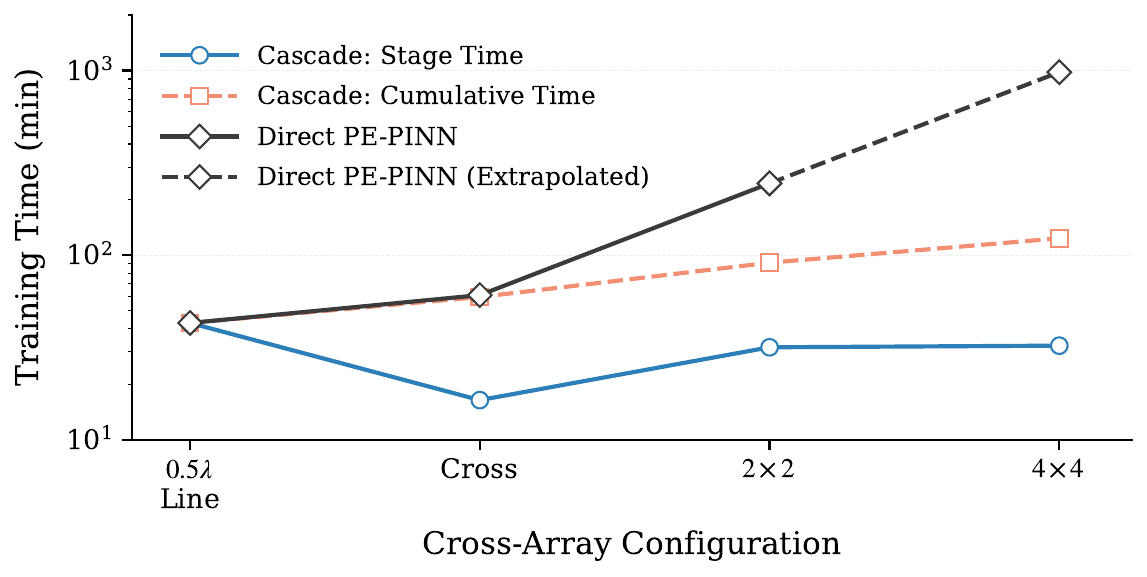}
    \caption{Training-time scaling for the hierarchical cross-array experiment. Stage time denotes the training cost of the current cascade level, while cumulative time includes all preceding stages from the elementary $0.5\lambda$ line-source model. The direct PE-PINN time for the $4\times4$ cross array is extrapolated based on the observed near-linear scaling with active kernel count.}
    \label{fig:cross_array_training_time}
\end{figure}

Figure~\ref{fig:cross_array_training_time} compares the training-time scaling of direct PE-PINN and PE-EK-PINN for the hierarchical cross-array experiments. The stage training time of PE-EK-PINN remains nearly constant after the elementary line-source model is obtained, increasing only from $16.4$ minutes for a single cross to approximately $32$ minutes for the $2\times2$ and $4\times4$ cross arrays. In contrast, the direct PE-PINN training time increases from $60.8$ minutes for a single cross to $245.1$ minutes for the $2\times2$ array as the number of active primitive kernels increases from 8 to 32. Based on this near-linear scaling, the direct training time for the $4\times4$ array is extrapolated to approximately 980 minutes (16.3 hours). Even when all preceding cascade stages are included, the cumulative PE-EK-PINN time for the $2\times2$ array is only $91.2$ minutes. These results further demonstrate that hierarchical kernel reuse substantially reduces the growth of training cost as the source system is enlarged.

\end{document}